# HCGMNET: A HIERARCHICAL CHANGE GUIDING MAP NETWORK FOR CHANGE DETECTION

*Chengxi Han[1], Chen Wu[1*], Bo Du[2]*

[1]State Key Laboratory of Information Engineering in Surveying, Mapping, and Remote Sensing, Wuhan University, Wuhan 430079, P. R. China
[2]School of Computer, Wuhan University, Wuhan 430072, P. R. China

## ABSTRACT

Very-high-resolution (VHR) remote sensing (RS) image change detection (CD) has been a challenging task for its very rich spatial information and sample imbalance problem. In this paper, we have proposed a hierarchical change guiding map network (HCGMNet) for change detection. The model uses hierarchical convolution operations to extract multi-scale features, continuously merges multi-scale features layer by layer to improve the expression of global and local information, and guides the model to gradually refine edge features and comprehensive performance by a change guide module (CGM), which is a self-attention with changing guide map. Extensive experiments on two CD datasets show that the proposed HCGMNet architecture achieves better CD performance than existing state-of-the-art (SOTA) CD methods.

## 1. INTRODUCTION

Change Detection (CD) is the process of identifying differences in the state of an object or phenomenon by observing it at different times [1].CD has a wide application in remote sensing interpretation tasks, including land use land cover analysis [2], urban extension studies [3], environmental monitoring [4], and disaster assessment [5].

The evaluation of the performance of a CD model mainly depends on the ability to extract the relevant changes and suppress the irrelevant changes, such as seasonal, illumination, environmental, and other interference changes. Many traditional pixel-level methods like change vector analysis (CVA), multivariate alteration detection (MAD), principal component analysis (PCA), and slow feature analysis (SFA), post-classification comparison (PCC), primarily depend on original image information and handcrafted features, which is easy to produce salt and pepper noise and irrelevant changes on Very-high-resolution (VHR) remote sensing (RS) images with more fine details and rich texture features.

Deep learning methods have made a remarkable performance on CD. Some models are proposed by stacked convolutional layers, such as FC-EF [6], FC-Siam-conc [6], and FC-Siam-diff [6], which can extract powerful discriminative features compared with traditional methods. However, most of them have a lot of holes in and error detection. Subsequently, researchers use dilated convolutions and attention mechanisms to enlarge the receptive field to get the details of global information, including STANet [7], SNUNet [8], and MSPSNet [9]. But, they still struggle to handle marginal details and omissions. Recently, transformers are widely used in the RS field including CD. BIT [10], Change Former [11], and RSP-BIT [12] are proposed to get effective receptive field and competitive performance. However, most of the transformers-based models consume a large amount of calculation and slow training speed, which limits their application and further study for researchers.

To better solve these problems, we propose an HCGMNet to use hierarchical convolution operations to extract multi-scale features, continuously merge multi-scale features layer by layer to improve the expression of global and local information, and guide the model to gradually refine edge features and comprehensive performance by **changing guide map**.

## 2. METHODOLOGY

As shown in Fig.1, HCGMNet is a hierarchical network which contains two parts: a multi-scale feature extractor to get coarse features of two input images and three Changing Guide Modules (CGM) to fuse multi-level fine features and produce the prediction.

### 2.1. Hierarchical multi-scale feature extractor

First, we use the VGG-16 network with batch normalization [13] as the backbone to extract the coarse features of the bi-temporal image. Five VGG16_BN Blocks denote the operation of vgg16_bn from the layer of 0-5,5-12,12-22,22-32,32-42. Subsequently, the hierarchical features are concatenated four times respectively. The convolutional block which contains the operation of the convolutional layer, batch normalization and ReLU, is used after four concatenations to refine features. At the same time, three of the four hierarchical features are up-sampled to a certain size and then the four features are concatenated together, which leads to a fusion of multi-scale features.

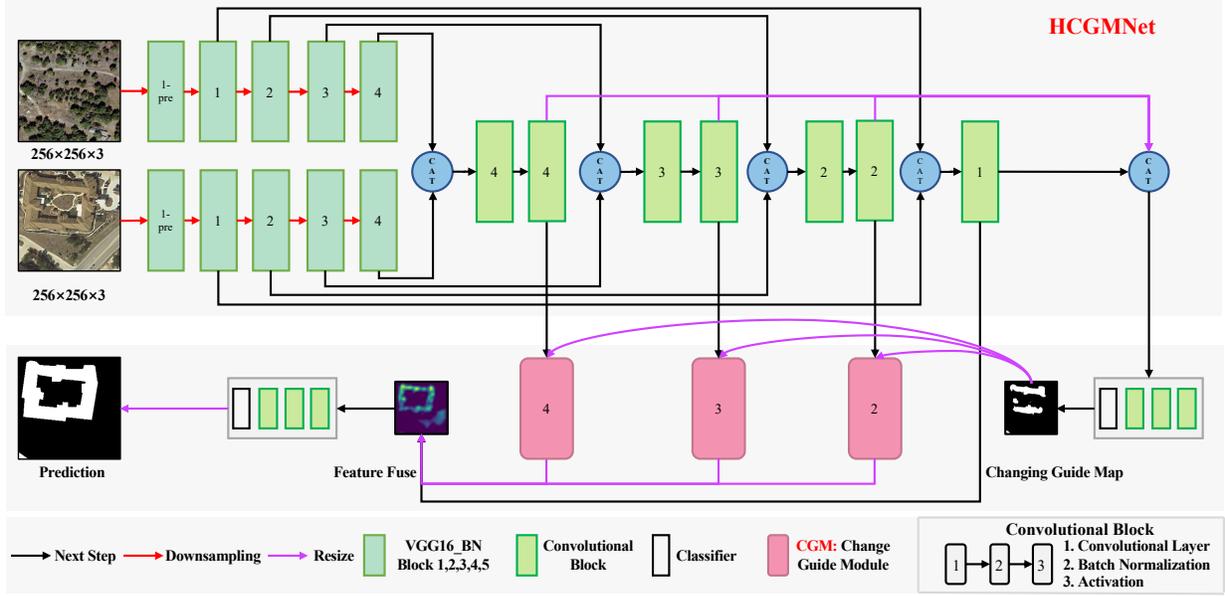

**Fig. 1.** Illustration of our proposed HCGMNet model.

## 2.2. Change Guide Module (CGM)

The semantic feature of HCGMNet is mainly realized from coarse to fine through CGM. As shown in Fig.2, CGM is a self-attention which can gradually fuse the feature of a guide map to extract the main change area. The original self-attention [14] is described as:

$$Attention(Q, K, V) = Softmax\left(\frac{QK^T}{\sqrt{d_{head}}}\right)V \qquad (1)$$

where $Q$, $K$, and $V$ denote Query, Key, and Value, respectively. But it's very computationally intensive because the dimensions of $Q$, $K$, and $V$ are all $HW \times C$, where $H$, $W$, and $C$ denote the size of the input image. Therefore, we reduce the parameters to improve the efficiency of computation. The channel of $Q$, $K$, and $V$ are compressed to an eighth of their original size. Through the feature fuse of Fig.2, it can be observed that the feature of the building can be clearly obtained from any one of the 512 layers' feature blocks. Notably, we propose that the effect of improving the expression of semantic features through a guide map is significant.

## 2.2. Loss Function

CD is an unbalanced binary classification task, thus we use cross-entropy and dice loss to reduce this imbalance. The dice loss function can be described as:

$$DiceLoss = 1 - \frac{2\sum_{i=1}^{N} y_i \hat{y}_i}{\sum_{i=1}^{N} y_i + \sum_{i=1}^{N} \hat{y}_i} \qquad (2)$$

where $y_i$ and $\hat{y}_i$ respectively represent the label value and predicted value of pixel $i$. $N$ is the total number of pixels, which is equal to the number of pixels in a single image multiplied by the batch size.

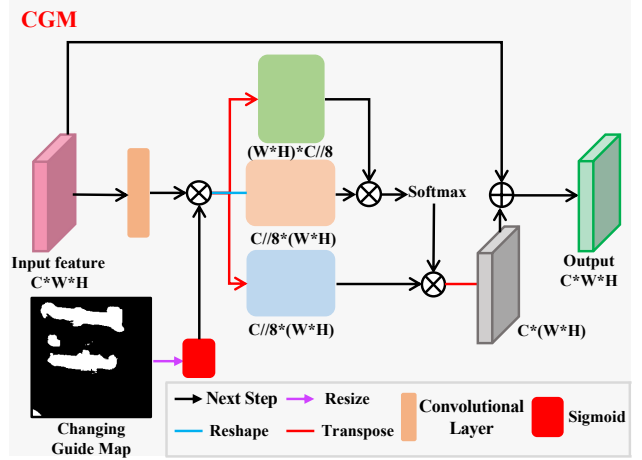

**Fig. 2.** Illustration of our CGM.

## 3. EXPERIMENTAL SETUP

### 3.1. Datasets

We test our proposed model in two publically available CD datasets, namely WHU-CD [15] and LEVIR-CD [7]. WHU-CD is a public remote sensing Building Change Detection dataset with 1260 training images and 690 images. We cropped the default image patch pairs into sizes of 256 × 256 with no overlap and obtained a dataset including 4536/504/2760 pairs of patches for training/validation/testing respectively. LEVIR-CD is a public large-scale remote sensing Building Change Detection dataset consisting of 637 VHR image patch pairs with a size of 1024 × 1024 pixels.

We cropped the default image patch pairs into sizes of 256 × 256 with no overlap and make train/val/test sets of samples 7120/1024/2048.

### 3.2. Implementation Details

We implement our models on PyTorch and use a single NVIDIA RTX 3090 GPU. We adopt the AdamW optimizer with a weight decay of 0.0025 and a learning rate of 5e-4 to minimize the loss. we set the batch size to 8 and the epoch number to 50.

### 3.3. Performance Metrics

For a more intuitive comparison, we use F1-score (F1), Precision (Pre.), Recall (Rec.), Overall Accuracy (OA), and Intersection over Union (IoU) scores to compare the performance of our model with SOTA methods. All of these metrics are employed by comparing the ground truth and prediction maps, and can be specifically calculated as follows:

$$F1 = \frac{2}{Pre.^{-1} + Rec.^{-1}} \quad (3)$$
$$Pre. = TP/(TP + FP) \quad (4)$$
$$Rec. = TP/(TP + FN) \quad (5)$$
$$OA = (TP + TN)/(TP + TN + FN + FP) \quad (6)$$
$$IoU = TP/(TP + FN + FP) \quad (7)$$

## 4. RESULTS AND DISCUSSION

In this section, we compare the CD performance of our HCGMNet with existing SOTA methods:
- **FC-EF [6]**: A Fully Convolutional Early Fusion network based on the U-Net structure.
- **FC-Siam-conc [6]**: A Fully Convolutional Siamese-Concatenation model.
- **FC-Siam-diff [6]**: A Fully Convolutional Siamese-Difference model.
- **STANet [7]**: A Siamese-based spatial-temporal attention neural network.
- **SNUNet [8]**: A densely connected Siamese network.
- **MSPSNet [9]**: A deep multiscale Siamese network.
- **BIT[10]**: A bitemporal image transformer network.
- **Change Former [11]**: A transformer-based Siamese network architecture.
- **RSP-BIT [12]**: A BIT model with RSP (Remote Sensing Pretraining).

TABLE I RESULTS OF COMPARISON WITH OTHER SOTA CHANGE DETECTION METHODS.

| Model | LEVIR-CD | | | | | WHU-CD | | | | |
|---|---|---|---|---|---|---|---|---|---|---|
| | F1 | Pre. | Rec. | OA | IoU | F1 | Pre. | Rec. | OA | IoU |
| FC-EF [6] | 61.52 | 73.31 | 53.00 | - | 44.43 | 58.05 | 76.49 | 46.77 | - | 40.89 |
| FC-Siam-conc [6] | 64.41 | **95.30** | 48.65 | - | 47.51 | 63.99 | 72.06 | 57.55 | - | 47.05 |
| FC-Siam-diff [6] | 89.00 | 91.76 | 86.40 | - | 80.18 | 86.31 | **89.63** | 83.22 | - | 75.91 |
| STANet-PAM [7] | 85.20 | 80.80 | 90.10 | 98.40 | 74.22 | 82.00 | 75.70 | 89.30 | 98.60 | 69.44 |
| SNUNet [8] | **89.97** | 91.31 | 88.67 | **98.99** | **81.77** | **87.76** | 87.84 | 87.68 | **99.13** | **78.19** |
| MSPSNet [9] | 89.67 | 90.75 | 88.61 | 98.96 | 81.27 | 86.49 | 87.84 | 85.17 | 99.05 | 76.19 |
| BIT [10] | 89.94 | 90.33 | **89.56** | 98.98 | 81.72 | 80.97 | 74.01 | **89.37** | 98.51 | 68.02 |
| Change Former [11] | **90.20** | 92.05 | 88.37 | **99.01** | **82.21** | 87.18 | **92.70** | 82.28 | **99.14** | **77.27** |
| RSP-BIT [12] | 89.71 | 92.00 | 87.53 | 98.98 | 81.34 | 78.50 | 69.93 | **89.45** | 98.26 | 64.60 |
| **HCGMNet (Ours)** | **91.77** | **92.96** | **90.61** | **99.18** | **84.79** | **92.08** | **93.93** | **90.31** | **99.45** | **85.33** |

* ALL VALUES ARE IN %. FOR CONVENIENCE: **BEST**, **2ND-BEST**, AND **3RD-BEST**.

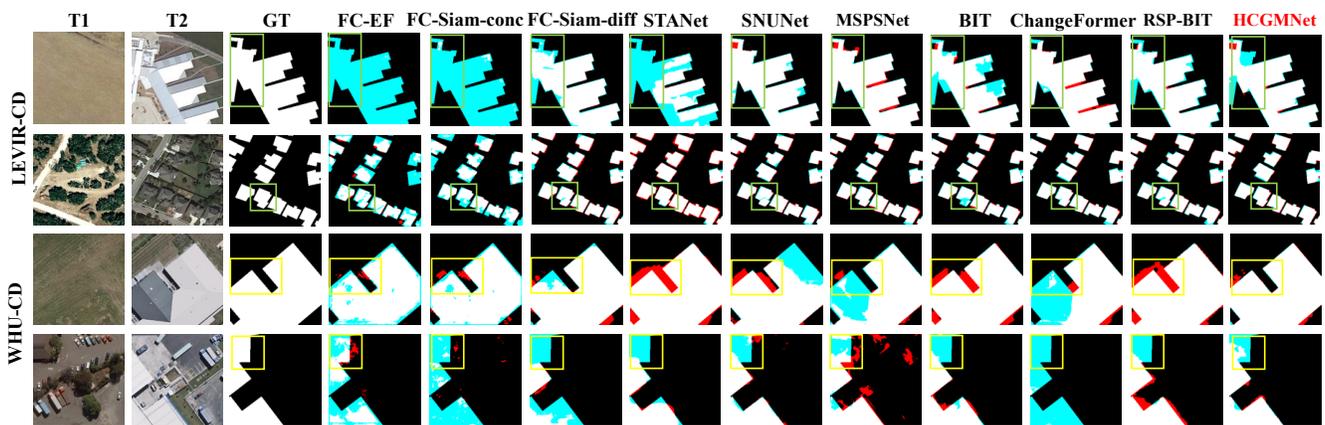

**Fig. 3.** Qualitative results of different methods on LEVIR-CD and WHU-CD. For convenience: TP (true positive, white), FP (false positive, red), TN (true negative, black), and FN (false negative, blue).

Tab. I show the results of comparison with other SOTA change detection methods on the test sets of LEVIR-CD and WHU-CD. As can be seen from the table, the proposed HCGMNet achieves better CD performance in terms of F1, OA, and IoU metrics. In addition, our HCGMNet improves previous SOTA in F1/ OA by 1.57/0.17% and in F1/ OA/ IoU by 4.32/0.31/1.33% for LEVIR-CD and WHU-CD, respectively. Especially in WHU-CD, HCGMNet gets first place on all metrics, which means our proposed method can work well. In this context, our proposed method can perform better than these models.

Fig. 3 compares the visualization of different SOTA change detection methods on test images from LEVIR-CD and WHU-CD. The red area means error detection (false positive) and the blue area means omissions (false negative). The stacked convolutional layers methods (FC-EF [6], FC-Siam-conc [6], and FC-Siam-diff [6]) have obvious omissions, the higher of the blue area, the more missed. Little omissions and error detection are detected in the dilated convolutions and attention mechanisms methods( STANet [7], SNUNet [8], and MSPSNet [9]). Compared with the previous two kinds of algorithms, the transformers-based methods( BIT [10], Change Former [11], and RSP-BIT [12]) have a better performance. From the green and yellow boxes, we can observe that our HCGMNet has lower FP and FN and can get better performance compared with other methods. In summary, our proposed HCGMNet has superiority in quantitative and qualitative comparisons.

## 5. CONCLUSION

In this paper, we propose a hierarchical change guiding map extraction network (HCGMNet), which uses hierarchical convolution operations to extract multi-scale features, continuously merges multi-scale features layer by layer to improve the expression of global and local information, and guides the model to gradually refine edge features and comprehensive performance by a change guide module (CGM). Two experiments are implemented on classical LEVIR-CD and WHU-CD datasets, and the qualitative and quantitative results validate the effectiveness and efficiency of the proposed method.

## 6. ACKNOWLEDGMENT

This work was supported in part by the National Key Research and Development Program of China under Grant 2022YFB3903302， and the National Natural Science Foundation of China under Grant T2122014 and 61971317.